\documentclass[runningheads]{llncs}
\usepackage{graphicx}
\usepackage{amssymb}
\usepackage{subcaption}
\usepackage{wrapfig}
\usepackage{booktabs}
\usepackage{hyperref,xcolor}
\usepackage[misc]{ifsym}
\begin{document}
\title{Cross-Dataset Adaptation for Instrument Classification in Cataract Surgery Videos}
%
%
\author{Jay N. Paranjape\inst{1,}\Letter \and
Shameema Sikder\inst{2,3} \and
Vishal M. Patel\inst{1} \and
S. Swaroop Vedula\inst{3}
}

\authorrunning{J. Paranjape et al.}
%
\institute{Department of Electrical and Computer Engineering, The Johns Hopkins University, Baltimore, USA \\
\email{jparanj1@jhu.edu}
\and
Wilmer Eye Institute, The Johns Hopkins University, Baltimore, USA \and Malone Center for Engineering in Healthcare, The Johns Hopkins University, Baltimore, USA
}

\maketitle              

\begin{abstract}
Surgical tool presence detection is an important part of the intra-operative and post-operative analysis of a surgery. State-of-the-art models, which perform this task well on a particular dataset, however, perform poorly when tested on another dataset. This occurs due to a significant domain shift between the datasets resulting from the use of different tools, sensors, data resolution etc. In this paper, we highlight this domain shift in the commonly performed cataract surgery and propose a novel end-to-end Unsupervised Domain Adaptation (UDA) method called the Barlow Adaptor that addresses the problem of distribution shift without requiring any labels from another domain. In addition, we introduce a novel loss called the Barlow Feature Alignment Loss (BFAL) which aligns features across different domains while reducing redundancy and the need for higher batch sizes, thus improving cross-dataset performance. The use of BFAL is a novel approach to address the challenge of domain shift in cataract surgery data. Extensive experiments are conducted on two cataract surgery datasets and it is shown that the proposed method outperforms the state-of-the-art UDA methods by 6\%. The code can be found at \url{https://github.com/JayParanjape/Barlow-Adaptor}

\end{abstract}
\section{Introduction}
\begin{figure}
\centering
\includegraphics[width=.75\textwidth]{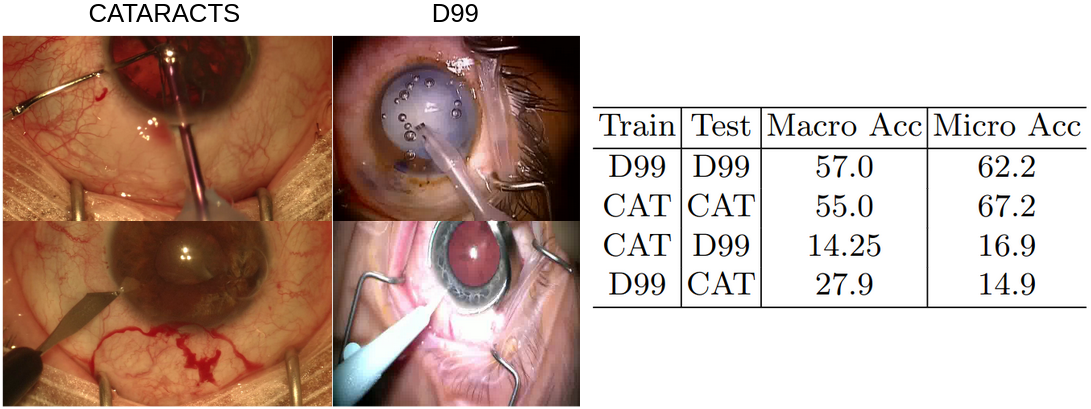}
\caption{Dataset shift between the CATARACTS dataset (CAT) \cite{cataracts} and D99 \cite{jhu1,jhu2} dataset. Results for models trained on one dataset and tested on another show a significant drop in performance.} \label{intro_pic}
\end{figure}

Surgical instrument identification and classification are critical to deliver several priorities in surgical data science \cite{vedula2022artificial}. Various deep learning methods have been developed to classify instruments in surgical videos using data routinely generated in institutions \cite{tool_detection_survey}. However, differences in image capture systems and protocols lead to nontrivial dataset shifts, causing a significant drop in performance of the deep learning methods when tested on new datasets \cite{VP_DA_survey}. Using cataract surgery as an example, Figure \ref{intro_pic} illustrates the drop in accuracy of existing methods to classify instruments when trained on one dataset and tested on another dataset \cite{endonet,deepphase}. Cataract surgery is one of the most common procedures \cite{cataract_common}, and methods to develop generalizable networks will enable clinically useful applications.

Domain adaptation methods aim to attempt to mitigate the drop in algorithm performance across domains \cite{VP_DA_survey}. Unsupervised Domain Adaptation (UDA) methods are particularly useful when the source dataset is labeled and the target dataset is unlabeled. In this paper, we describe a novel end-to-end UDA method, which we call the Barlow Adaptor, and its application for instrument classification in video images from cataract surgery. We define a novel loss function called the Barlow Feature Alignment Loss (BFAL) that aligns the features learnt by the model between the source and target domains, without requiring any labeled target data. It encourages the model to learn non-redundant features that are domain agnostic and thus tackles the problem of UDA. BFAL can be added as an add-on to existing methods with minimal code changes.
The contributions of our paper are threefold:
\begin{enumerate}
    \item We define a novel loss for feature alignment called BFAL that doesn't require large batch sizes and encourages learning non-redundant, domain agnostic features.
    \item We use BFAL to generate an end-to-end system called the Barlow Adaptor that performs UDA. We evaluate the effectiveness of this method and compare it with existing UDA methods for instrument classification in cataract surgery images.
    \item We motivate new research on methods for generalizable deep learning models for surgical instrument classification using cataract surgery as the test-bed. Our work proposes a solution to the problem of lack of generalizability of deep learning models that was identified in previous literature on cataract surgery instrument classification.
\end{enumerate}

\section{Related Work}
\noindent {\bf{Instrument Identification in Cataract Surgery Video Images.}} The motivation for instrument identification is its utility in downstream tasks such as activity localization and skill assessment  \cite{sr1,sr2,cataracts_dl}. The current state-of-the-art instrument identification method called Deep-Phase \cite{deepphase} uses a ResNet architecture to identify instruments and then to identify steps in the procedure. However, a recent study has shown that while these methods work well on one dataset, there is a significant drop in performance when tested on a different dataset \cite{da_cataracts}. Our analyses reiterate similar findings on drop in performance (Figure \ref{intro_pic}) and highlight the effect of domain shift between data from different institutions even for the same procedure.\\
\noindent {\bf{Unsupervised Domain Adaptation.}} UDA is a special case of domain adaptation, where a model has access to annotated training data from a source domain and unannotated data from a target domain \cite{VP_DA_survey}. Various methods have been proposed in the literature to perform UDA. One line of research involves aligning the feature distributions between the source and target domains. Maximum Mean Discrepancy (MMD) is commonly used as a distance metric between the source and target distributions \cite{mmd}. Other UDA methods use a convolutional neural network (CNN) to generate features and then use MMD as an additional loss to align distributions \cite{mmd1,mmd2,mmd3,mmd4,mmd5,mmd6}. While MMD is a first-order statistic, Deep CORAL \cite{deep_coral} penalizes the difference in the second-order covariance between the source and target distributions. Our method uses feature alignment by enforcing a stricter loss function during training.

Another line of research for UDA involves adversarial training. Domain Adaptive Neural Network (DANN) \cite{dann} involves a minimax game, in which one network minimizes the cross entropy loss for classification in the source domain, while the other maximizes the cross entropy loss for domain classification. Few recent methods generate pseudo labels on the target domain and then train the network on them. One such method is Source Hypothesis Transfer (SHOT) \cite{shot}, which performs source-free domain adaptation by further performing information maximization on the target domain predictions. While CNN-based methods are widely popular for UDA, there are also methods which make use of the recently proposed Vision Transformer (ViT) \cite{vit}, along with an ensemble of the above described UDA based losses. A recent approach called Cross Domain Transformer (CDTrans) uses cross-domain attention to produce pseudo labels for training that was evaluated in various datasets \cite{cdtrans}. Our proposed loss function is effective for both CNN and ViT-based backbones.

\section{Proposed Method}

In the UDA task, we are given \(n_s\) observations from the source domain $\mathcal{D}_S$. Each of these observations is in the form of a tuple \((x_s,y_s)\), where \(x_s\) denotes an image from the source training data and \(y_s\) denotes the corresponding label, which is the instrument index present in the image. In addition, we are given \(n_t\) observations from the target domain $\mathcal{D}_T$. Each of these can be represented by \({x_t}\), which represents the image from the target training data. However, there are no labels present for the target domain during training. The goal of UDA is to predict the labels \(y_t\) for the target domain data.\\

\noindent {\bf{Barlow Feature Alignment Loss (BFAL).}} We introduce a novel loss, which encourages features between the source and target to be similar to each other while reducing the redundancy between the learnt features. BFAL works on pairs of feature projections of the source and target. More specifically, let \(f_s \in \mathbb{R}^{BXD}\) and \(f_t \in \mathbb{R}^{BXD}\) be the features corresponding to the source and target domain, respectively. Here \(B\) represents the batch size and \(D\) represents the feature dimension. Similar to \cite{barlow}, we project these features into a \(P\) dimensional space using a fully connected layer called the Projector, followed by a batch normalization to whiten the projections. Let the resultant projections be denoted by \(p_s \in \mathbb{R}^{BXP}\) for the source and \(p_t \in \mathbb{R}^{BXP}\) for the target domains. Next, we compute the correlation matrix \(\mathbb{C}_1 \in \mathbb{R}^{PXP}\). Each element of \(\mathbb{C}_1\) is computed as follows
\setlength{\belowdisplayskip}{0pt} \setlength{\belowdisplayshortskip}{0pt}
\setlength{\abovedisplayskip}{0pt} \setlength{\abovedisplayshortskip}{0pt}
\begin{equation}\label{c1}
    \mathbb{C}_1^{ij} = \frac{\sum_{b=1}^{B} p_{s}^{bi} p_{t}^{bj}}{\sqrt{\sum_{b=1}^{B} (p_{s}^{bi})^2} \sqrt{\sum_{b=1}^{B} (p_{t}^{bj})^2}}.  
\end{equation}
Finally, the BFAL is computed using the L2 loss between the elements of \(\mathbb{C}_1\) and the identity matrix \(\mathbb{I}\) as follows
\begin{equation}\label{bfal}
    \mathbb{L}_{BFA} = \underbrace{\sum_{i=1}^{P} (1 - \mathbb{C}_{1}^{ii})^{2}}_{feature \: alignment} + \underbrace{\mu \sum_{i=1}^{P}\sum_{j \neq i} (\mathbb{C}_{1}^{ij})^2}_{redundancy \: reduction},
\end{equation}
where $\mu$ is a constant.
Intuitively, the first term of the loss function can be thought of as a feature alignment term since we push the diagonal elements in the covariance matrix towards 1. In other words, we encourage the feature projections between the source and target to be perfectly correlated. On the other hand, by pushing the off-diagonal elements to 0, we decorrelate different components of the projections. Hence, this term can be considered a redundancy reduction term, since we are pushing each feature vector component to be independent of one another. 

BFAL is inspired by a recent technique in self-supervised learning, called the Barlow Twins \cite{barlow}, where the authors show the effectiveness of such a formulation at lower batch sizes. In our experiments, we observe that even keeping a batch size of 16 gave good results over other existing methods. Furthermore, BFAL does not require large amounts of data to converge.\\

\begin{figure}
\centering
\includegraphics[width=\textwidth]{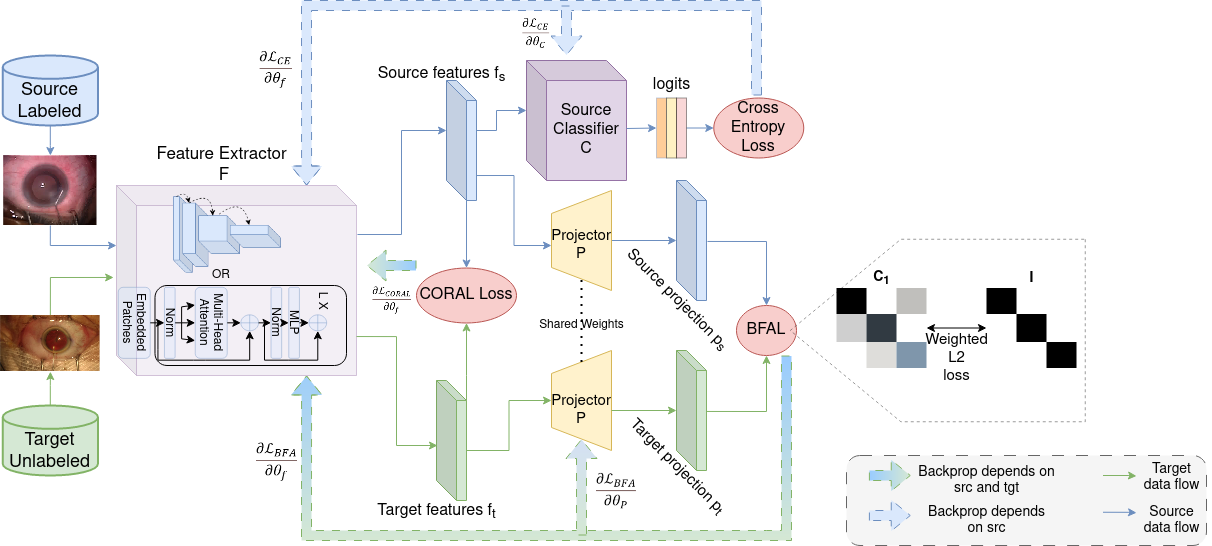}
\caption{Architecture corresponding to the Barlow Adaptor. Training occurs using pairs of images from the source and target domain. They are fed into the feature extractor, which generates features used for the CORAL loss. Further, a projector network $P$ projects the features into a $P$ dimensional space. These are then used to calculate the Barlow Feature Alignment Loss. One branch from the source features goes into the source classifier network that is used to compute the cross entropy loss with the labeled source data. [Backprop = backpropagation; src = source dataset, tgt = target dataset]} \label{ba_arch}
\end{figure}

\noindent {\bf{Barlow Adaptor.}} We propose an end-to-end method that utilizes data from the labeled source domain and the unlabeled target domain. The architecture corresponding to our method is shown in Figure \ref{ba_arch}.

There are two main sub-parts of the architecture - the Feature Extractor \(F\), and the Source Classifier \(C\). First, we divide the training images randomly into batches of pairs \(\{x_s,x_t\}\) and apply \(F\) on them, which gives us the features extracted from these sets of images. For the Feature Detector, we show the effectiveness of our novel loss using ViT and ResNet50 both of which have been pre-trained on ImageNet. The features obtained are denoted as \(f_s\) and \(f_t\) for the source and target domains, respectively. Next, we apply \(C\) on these features to get logits for the classification task. The source classifier is a feed forward neural network, which is initialized from scratch. These logits are used, along with the source labels \(y_s\) to compute the source cross entropy loss as $\mathbb{L}_{CE} = \frac{-1}{B}\sum_{b=1}^B {\sum_{m=1}^My_{s}^{bm}\log(p_{s}^{bm})},$

where \(M\) represents the number of classes, \(B\) represents the total mini-batches, while \(m\) and \(b\) represent their respective indices. 

The features \(f_s\) and \(f_t\) are further used to compute the Correlation Alignment(CORAL) loss and the BFAL, which enforce the feature extractor to align its weights so as to learn features that are domain agnostic as well as non-redundant. The BFAL is calculated as mentioned in the previous subsection. The CORAL loss is computed as depicted in Equation \ref{coral}, following the UDA method Deep CORAL \cite{deep_coral}. While the BFAL focuses on reducing redundancy, CORAL works by aligning the distributions between the source and target domain data. This is achieved by taking the difference between the covariance matrices of the source and target features - \(f_s\) and \(f_t\) respectively. The final loss is the weighted sum of the three individual losses as follows:
\begin{equation}\label{final}
    \mathbb{L}_{final} = \mathbb{L}_{CE} + \lambda(\mathbb{L}_{CORAL} + \mathbb{L}_{BFA}), 
\end{equation}
where 
\begin{equation}\label{coral}
    \mathbb{L}_{CORAL} = \frac{1}{4D^2} \| \mathbb{C}_s - \mathbb{C}_t \|_F^2,  \;\;\;  \mathbb{C}_s = \frac{1}{B-1} (f_s^T f_s) - \frac{1}{B}(\textbf{1}^T f_{s})^T (\textbf{1}^T f_{s}),
\end{equation}

\begin{equation}\label{ct}
    \mathbb{C}_t = \frac{1}{B-1} (f_t^T f_t) - \frac{1}{B}(\textbf{1}^T f_{t})^T (\textbf{1}^T f_{t}).
\end{equation}

Each of these three losses plays a different role in the UDA task. The cross entropy loss encourages the model to learn discriminative features between images with different instruments. The CORAL loss pushes the features between the source and target towards having a similar distribution. Finally, the BFAL tries to make the features between the source and the target non-redundant and same. BFAL is a stricter loss than CORAL as it forces features to not only have the same distribution but also be equal. Further, it also differs from CORAL in learning independent features as it explicitly penalizes non-zero non-diagonal entries in the correlation matrix. While using BFAL alone gives good results, using it in addition to CORAL gives slightly better results empirically. We note these observations in our ablation studies. Between the cross entropy loss and the BFAL, an adversarial game is played where the former makes the features more discriminative and the latter tries to make them equal. The optimal features thus learnt are different in aspects required to identify instruments but are equal for any domain-related aspect. This property of the Barlow Adaptor is especially useful for surgical domains where the background has similar characteristics for most of the images within a domain. For example, for cataract surgery images, the position of the pupil or the presence of blood during the usage of certain instruments might be used by the model for classification along with the instrument features. These features depend highly upon the surgical procedures and the skill of the surgeon, thus making them highly domain-specific and possibly unavailable in the target domain. Using BFAL during training attempts to prevent the model from learning such features. 

\section{Experiments and Results}
We evaluate the proposed UDA method for the task of instrument classification using two cataract surgery image datasets. In our experiments, one dataset is used as the source domain and the other is used as the target domain. We use micro and macro accuracies as our evaluation metrics. Micro accuracy denotes the number of correctly classified observations divided by the total number of observations. In contrast, macro accuracy denotes the average of the classwise accuracies and is effective in evaluating classes with less number of samples.\\

\noindent {\bf{Datasets.}} The first dataset we use is CATARACTS \cite{cataracts}, which consists of 50 videos with framewise annotations available for 21 surgical instruments. The dataset is divided into 25 training videos and 25 testing videos. We separate 5 videos from the training set and use them as the validation set for our experiments. The second dataset is called D99 in this work \cite{jhu1,jhu2}, which consists of 105 videos of cataract surgery with annotations for 25 surgical instruments. Of the 105 videos, we use 65 videos for training, 10 for validation and 30 for testing. We observe a significant distribution shift between the two datasets as seen in Figure \ref{intro_pic}. This is caused by several factors such as lighting, camera resolution, and differences in instruments used for the same steps. For our experiments in this work, we use 14 classes of instruments that are common to both datasets. Table \ref{tab_map} shows a mapping of instruments between the two datasets. For each dataset, we normalize the images using the means and standard deviations calculated from the respective training images. In addition, we resize all images to $224\times224$ size and apply random horizontal flipping with a probability of 0.5 before passing them to the model.

\begin{table}
\caption{Mapping of surgical tools between CATARACTS(L) and D99(R)}\label{tab_map}
\resizebox{\textwidth}{!}{\begin{tabular}{|c|c|||c|c|}
\hline
CATARACTS &  D99 & CATARACTS & D99\\
\hline
Secondary Incision Knife & Paracentesis Blade & Bonn Forceps & 0.12 Forceps\\
Charleux Cannula & Anterior Chamber Cannula & Irrigation & Irrigation\\
Capsulorhexis Forceps & Utrata Forceps & Cotton & Weckcell Sponge\\
Hydrodissection Cannula & Hydrodissection Cannula & Implant Injector & IOL Injector\\
Phacoemulsifier Handpiece & Phaco Handpiece & Suture Needle & Suture\\
Capsulorhexis Cystotome & Cystotome & Needle Holder & Needle Driver\\
Primary Incision Knife & Keratome & Micromanipulator & Chopper\\
\hline
\end{tabular}}
\end{table}


\noindent {\bf{Experimental Setup.}}
We train the Barlow Adaptor for multi-class classification with the above-mentioned 14 classes in Pytorch. For the Resnet50 backbone, we use weights pretrained on Imagenet \cite{imagenet} for initialization. For the ViT backbone, we use the base-224 class of weights from the TIMM library \cite{timm}. The Source Classifier $C$ and the Projector $P$ are randomly initialized. We use the validation sets to select the hyperparameters for the models. Based on these empirical results, we choose \(\lambda\) from Equation \ref{final} to be 0.001 and \(\mu\) from Equation \ref{bfal} to be 0.0039. We use SGD as the optimizer with momentum of 0.9 and a batch size of 16. We start the training with a learning rate of 0.001 and reduce it by a factor of 0.33 every 20 epochs. The entire setup is trained with a single NVIDIA Quatro RTX 8000 GPU. We use the same set of hyperparameters for the CNN and ViT backbones in both datasets.

\begin{table}
\centering
\caption{Macro and micro accuracies for cross domain tool classification. Here, source-only denotes models that have only been trained on one domain and tested on the other. Similarly, target-only denotes models that have been trained on the test domain and thus act as an upper bound. Deep CORAL \cite{deep_coral} is similar to using CORAL with ResNet backbone, so we don't list the latter separately. Here, CAT represents the CATARACTS dataset.}
\label{tab_vit}
\begin{tabular}{|c|c|c|c|c|}
\hline
&\multicolumn{2}{|c|}{ D99 $\rightarrow$ CAT} & \multicolumn{2}{|c|}{CAT $\rightarrow$ D99}\\
\hline
Method &  Macro Acc & Micro Acc & Macro Acc & Micro Acc \\
\hline
Source Only (ResNet50 backbone) & 27.9\% & 14.9\% & 14.25\% & 16.9\%\\
MMD with ResNet50 backbone\cite{mmd} & 32.2\% & 15.9\% & 20.6\% & 24.3\% \\
\hline
Source Only (ViT backbone) & 30.43\% & 14.14\% & 13.99\% & 17.11\%\\
MMD with ViT backbone\cite{mmd} & 31.32\% & 13.81\% & 16.42\% & 20\% \\
CORAL with ViT backbone\cite{deep_coral} & 28.7\% & 16.5\% & 15.38\% & 18.5\\
\hline
DANN\cite{dann} & 22.4\% & 11.6\% & 16.7\% & 19.5\%\\
Deep CORAL\cite{deep_coral} & 32.8\% & 14\% & 18.6\% & 22\\
CDTrans\cite{cdtrans} & 29.1\% & 14.7\% & 20.9\% & 24.7\%\\
\hline
Barlow Adaptor with ResNet50 (Ours) & \textbf{35.1\%} & \textbf{17.1\%} & \textbf{24.62\%} & \textbf{28.13\%}\\
Barlow Adaptor with ViT (Ours) & 31.91\% & 12.81\% & 17.35\% & 20.8\%\\
\hline
Target Only (ResNet50) & 55\% & 67.2\% & 57\% & 62.2\% \\
Target Only (ViT) & 49.80\% & 66.33\% & 56.43\% & 60.46\% \\
\hline
\end{tabular}
\end{table}

\noindent {\bf{Results.}} Table \ref{tab_vit} shows results comparing the performance of the Barlow Adaptor with recent UDA methods. We highlight the effect of domain shift by comparing the source-only models and the target-only models, where we observe a significant drop of 27\% and 43\% in macro accuracy for the CATARACTS dataset and the D99 dataset, respectively. Using the Barlow Adaptor, we observe an increase in macro accuracy by 7.2\% over the source only model. Similarly, we observe an increase in macro accuracy of 9\% with the Barlow Adaptor when the source is CATARACTS and the target is the D99 dataset compared with the source only model. Furthermore, estimates of macro and micro accuracy are larger with the Barlow Adaptor than those with other existing methods. Finally, improved accuracy with the Barlow Adaptor is seen with both ResNet and ViT backbones. 

\noindent {\bf{Ablation Study.}} We tested the performance gain due to each part of the Barlow Adaptor. Specifically, the Barlow Adaptor has CORAL loss and BFAL as its two major feature alignment losses. We remove one component at a time and observe a decrease in performance with both ResNet and ViT backbones (Table \ref{tab_vit_abl}). This shows that each loss has a part to play in domain adaptation. Further ablations are included in the supplementary material.

\begin{table}
\centering
\caption{Findings from ablation studies to evaluate the Barlow Adaptor. Here, Source Only is the case where neither CORAL nor BFAL is used. We use Macro Accuracy for comparison. Here, CAT represents the CATARACTS dataset.}\label{tab_vit_abl}
\begin{tabular}{|c|c|c|c|c|}
\hline
& \multicolumn{2}{|c|}{ViT Feature Extractor} & \multicolumn{2}{|c|}{ResNet50 Feature Extractor}\\
\hline
Method &  D99 $\rightarrow$ CAT & CAT $\rightarrow$ D99 &  D99 $\rightarrow$ CAT & CAT $\rightarrow$ D99\\
\hline
Source Only(\(\mathbb{L}_{CE}\)) & 30.43\% & 16.7\% & 27.9\% & 14.9\%\\
Only CORAL(\(\mathbb{L}_{CORAL}\)) & 28.7\% & 15.38\% & 32.8\% & 18.6\%\\
Only BFAL(\(\mathbb{L}_{BFA}\)) & 29.8\% & 17.01\% & 32.3\% & 24.46\%\\
Barlow Adaptor(Eq \ref{final}) & \textbf{32.1}\% & \textbf{17.35}\% & \textbf{35.1\%} & \textbf{24.62}\%\\
\hline
\end{tabular}
\end{table}

\section{Conclusion}
Domain shift between datasets of cataract surgery images limits generalizability of deep learning methods for surgical instrument classification. We address this limitation using an end-to-end UDA method called the Barlow Adaptor. As part of this method, we introduce a novel loss function for feature alignment called the BFAL. Our evaluation of the method shows larger improvements in classification performance compared with other state-of-the-art methods for UDA. BFAL is an independent module and can be readily integrated into other methods as well. BFAL can be easily extended to other network layers and architectures as it only takes pairs of features as inputs.
\section{Acknowledgement}
This research was supported by a grant from the National Institutes of Health, USA; R01EY033065. The content is solely the responsibility of the authors and does not necessarily represent the official views of the National Institutes of Health. 

%
%
%
%
\bibliographystyle{splncs04}
\bibliography{bibliography}

\end{document}


\title{Supplementary - Cross-Dataset Adaptation for Instrument Classification in Cataract Surgery Videos
}

\author{Jay N. Paranjape\inst{1,}\Letter \and
Shameema Sikder\inst{2,3} \and
Vishal M. Patel\inst{1} \and
S. Swaroop Vedula\inst{3}
}
\authorrunning{J. Paranjape et al.}
%
\institute{Department of Electrical and Computer Engineering, The Johns Hopkins University, Baltimore, USA \\
\email{jparanj1@jhu.edu}
\and
Wilmer Eye Institute, The Johns Hopkins University, Baltimore, USA \and Malone Center for Engineering in Healthcare, The Johns Hopkins University, Baltimore, USA
}
\maketitle            

\begin{figure}
\begin{subfigure}{0.5\textwidth}
\includegraphics[width=0.9\linewidth]{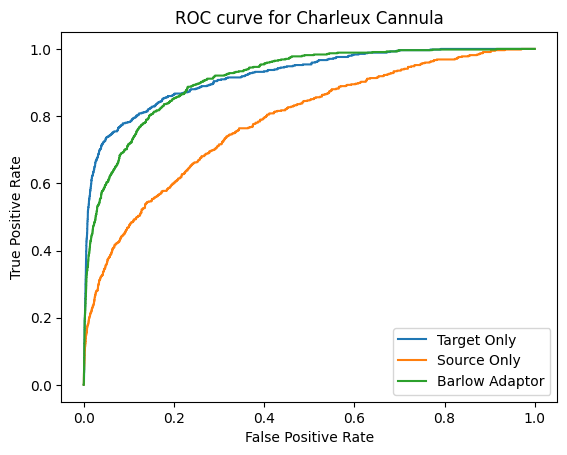} 
\label{fig:ac_cannula}
\end{subfigure}
\begin{subfigure}{0.5\textwidth}
\includegraphics[width=0.9\linewidth]{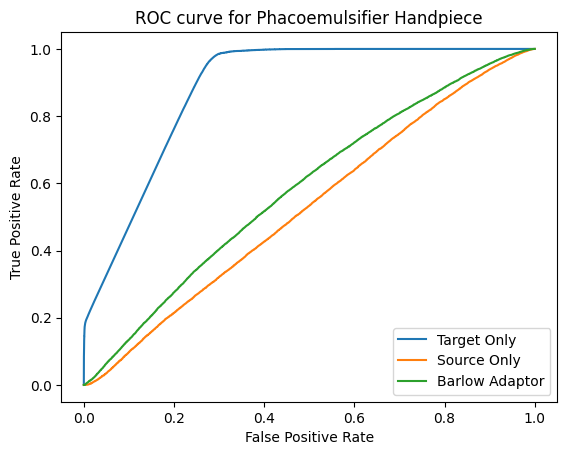}
\label{fig:phaco_roc}
\end{subfigure}
\caption{ROC curve for the Barlow Adaptor matches the target-only model for the AC Cannula. This shows a successful domain transfer. However, its not able to perform well for classifying the Phacoemulsifier handpiece.}
\label{rocs}
\end{figure}

\begin{figure}
\centering
\includegraphics[width=.5\textwidth]{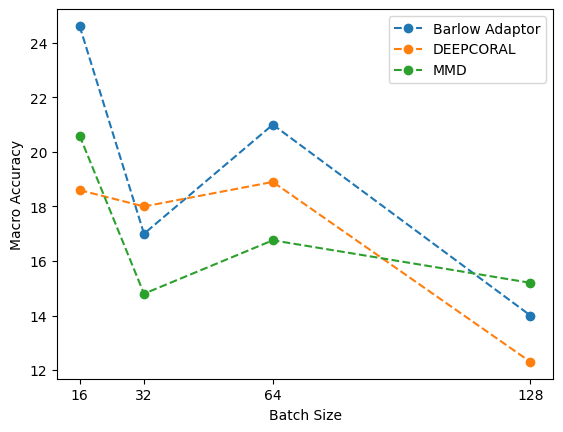}
\caption{Effect of batch size on macro accuracy for the ResNet50 feature extractor and CATARACTS to D99 transfer. The Barlow Adaptor works better for a low batch size of 16 over its counterparts.}
\label{bs_abl}
\end{figure}

To understand the instrument-wise performance of our model, we follow the report on the CATARACTS challenge and compare its area under the receiver operating characteristic curve (AUROC) with that of the target-only and source-only models. For the CATARACTS dataset, we observe a macro AUROC of 0.96 for the target-only model. Owing to a significant domain shift, this drops to 0.78 for the source-only model. With the Barlow Adaptor, we are able to increase it to 0.79. Similarly, for the D99 dataset, the macro AUROC drops from 0.92 for the target-only model to 0.66 for the source-only model. We are able to improve it to 0.73. Fig \ref{rocs} shows the receiver operating characteristic (ROC) curves for the AC Cannula and PhacoEmulsifier Handpiece. The Barlow Adaptor was able to transfer knowledge well for the former. However, there is a scope for improvement in classifying the latter. 
\begin{wrapfigure}{r}{0.5\textwidth}
\includegraphics[width=.48\textwidth]{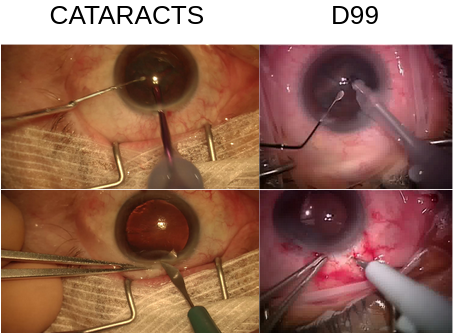}
\caption{(Top row) Phacoemulsifier Handpiece used with Micromanipulator. (Bottom Row) 0.12 Forceps used with the Secondary Incision Knife.}
\label{two_ins}
\end{wrapfigure}
On manual inspection of the dataset, we found  that a major reason for this is the existence of multiple instruments being used together commonly. In the above example, we observe that the Phacoemulsifier Handpiece is commonly used along with the Micromanipulator, thus reducing the scores for these instruments. Similarly, 0.12 Forceps is used commonly along with the Secondary Incision Knife. Figure \ref{two_ins} shows this for both datasets. Future work will include handling such multi-label cases.
Further, we show that our method works well with lower batch sizes, which makes it less expensive computationally. Figure \ref{bs_abl} shows a comparison between different methods for increasing batch sizes.

We tabulate the classwise accuracies for each of the instruments in tables \ref{tab_all} 
Instruments that are commonly used with other instruments and those that are rare in the training set show low accuracies.

\begin{table}
\centering
\caption{Instrument Wise Accuracies for Cross-Domain Tool Classification. (1) Secondary Incision Knife (2) Bonn Forceps (3) Charleux Cannula (4) Primary Incision Knife (5) Capsulorhexis forceps (6) Capsulorhexis Cystotome (7) Hydrodissection Cannula (8) PhacoEmulsifier Handpiece (9)Implant Injector (10) Suture Needle (11) Cotton (12) Needle Holder (13) Irrigation (14) Micromanipulator. MMD = Maximum Mean Discrepancy; ViT = Vision Transformer; CORAL = Correlation Alignment; DANN = Domain Adaptation Neural Network; CDTrans = Cross Domain Transformer.}
\label{tab_all}

\resizebox{\textwidth}{!}{\begin{tabular}{|c|c|c|c|c|c|c|c|c|c|c|c|c|c|c|c|}
\hline
& \multicolumn{15}{|c|}{Pvt $\rightarrow$ CAT}\\
\hline
Method & 1 & 2 & 3 & 4 & 5 & 6 & 7 & 8 & 9 & 10 & 11 & 12 & 13 & 14 &Average\\
\hline
Source Only(ResNet50 backbone) & 14.25\% & 10.98\% & 39.85\% & 40.13\% & 64.90\% & 24.29\% & 58.67\% & 1.16\% & 71.15\% & 0\% & 27.27\% & 13.6\% & 0.59\% & 23.96\% & 27.9\% \\
MMD with ResNet50 backbone & 0.24\% & 35.94\% & 77.12\% & 30.97\% & 53.01\% & 30.34\% & 36.95\% & 1.95\% & 60.66\% & 3.11\% & 63.64\% & 30.4\% & 1.66\% & 25.45\% & 32.24\% \\
\hline
Source Only(ViT backbone) & 2.9\% & 53.23\% & 67.34\% & 19.9\% & 54.48\% & 8.88\% & 40.03\% & 0.43\% & 72.42\% & 0.78\% & 63.64\% & 15.2\% & 1.42\% & 25.38\% & 30.43\% \\
MMD with ViT backbone & 3.62\% & 51.13\% & 54.80\% & 23.01\% & 56.98\% & 22.80\% & 36.22\% & 0.86\% & 68.87\% & 0.39\% & 63.64\% & 33.60\% & 2.30\% & 20.25\% & 31.32\% \\
CORAL with ViT backbone & 0.48\% & 59.55\% & 18.63\% & 18.69\% & 60.35\% & 27.68\% & 50.65\% & 0.31\% & 68.78\% & 0\% & 54.55\% & 18.4\% & 0.2\% & 24.63\% & 28.78\% \\
\hline
DANN & 14.01\% & 21.20\% & 44.83\% & 15.57\% & 30.54\% & 26.82\% & 24.55\% & 1.51\% & 46.44\% & 0.78\% & 45.45\% & 21.60\% & 9.96\% & 10.74\% & 22.43\% \\
Deep CORAL & 13.04\% & 24.81\% & 94.46\% & 32.87\% & 42.0\% & 48.29\% & 43.27\% & 1.38\% & 52.45\% & 1.56\% & 72.73\% & 15.2\% & 1.96\% & 15.9\% & 32.85\% \\
CDTrans & 0\% & 1.05\% & 5.17\% & 64.53\% & 68.72\% & 78.57\% & 17.75\% & 1.36\% & 30.12\% & 5.45\% & 90.91\% & 26.40\% & 0.01\% & 17.51\% & 29.1\% \\
\hline
Barlow Adaptor with ResNet50 (Ours) & 21.5\% & 26.17\% & 69.00\% & 42.56\% & 49.19\% & 46.18\% & 38.98\% & 6.05\% & 63.87\% & 3.11\% & 81.82\% & 18.4\% & 3.52\% & 21.25\% & 35.11\% \\
Barlow Adaptor with ViT (Ours) & 0.97\% & 40.45\% & 73.8\% & 41.18\% & 55.51\% & 23.74\% & 27.23\% & 0.71\% & 87.9\% & 0\% & 63.64\% & 14.4\% & 0.49\% & 16.68\% & 31.91\% \\
\hline
Target Only(ResNet50) & 84.06\% & 28.72\% & 6.64\% & 88.06\% & 72.83\% & 89.38\% & 89.87\% & 44.15\% & 87.39\% & 28.40\% & 0\% & 0.8\% & 89.84\% & 61.65\% & 55.13\% \\
Target Only(ViT) & 50.97\% & 53.08\% & 9.23\% & 70.76\% & 49.49\% & 85.33\% & 86.55\% & 86.55\% &90.02\% & 0.78\% & 0\% & 0\% & 87.52\% & 26.91\% & 49.80\% \\
\hline
\end{tabular}}
\end{table}